\documentclass[letterpaper]{article} 
\usepackage[preprint]{aaai2027}
\usepackage[hyphens]{url} 
\usepackage{graphicx} 
\usepackage{natbib} 
\usepackage{caption} 
\usepackage{amsmath}
\usepackage{amsfonts}
\usepackage{booktabs}
\usepackage{colortbl}
\usepackage{marvosym}
\usepackage{multirow}

\newcommand{\jr}[1]{#1}
\newcommand{\jrc}[1]{}
\newcommand{\clh}[1]{}

\newcommand{\eg}{\emph{e.g.}}
\newcommand{\cqf}[1]{}

\newcommand{\correspondingicon}{\textsuperscript{\Letter}}

\title{USR-Drive: Unified Driving Scene Representation via Joint Denoising of 3D Gaussians and Boxes}
\author{Li-Heng Chen\textsuperscript{\rm 1,}\textsuperscript{\rm 2} \qquad Haokai Pang\textsuperscript{\rm 2} \qquad Chengye Su\textsuperscript{\rm 1} \qquad
Jiarun Liu\textsuperscript{\rm 1} \qquad Qifeng Chen\textsuperscript{\rm 1} \\ Ziqian Ni\textsuperscript{\rm 1} \qquad
Jianxin Huang\textsuperscript{\rm 2} \qquad Shi-Sheng Huang\textsuperscript{\rm 3} \qquad Hongbo Fu\textsuperscript{\rm 2}\correspondingicon \qquad Sheng Yang\textsuperscript{\rm 1}\correspondingicon}
\affiliations{\textsuperscript{\rm 1}NIO \qquad \textsuperscript{\rm 2}HKUST \qquad \textsuperscript{\rm 3}Beijing Normal University}

\begin{document}


\twocolumn[{%
  \renewcommand\twocolumn[1][]{#1}
  \maketitle

  \begin{center}
    \captionsetup{type=figure}
    \includegraphics[width=\textwidth]{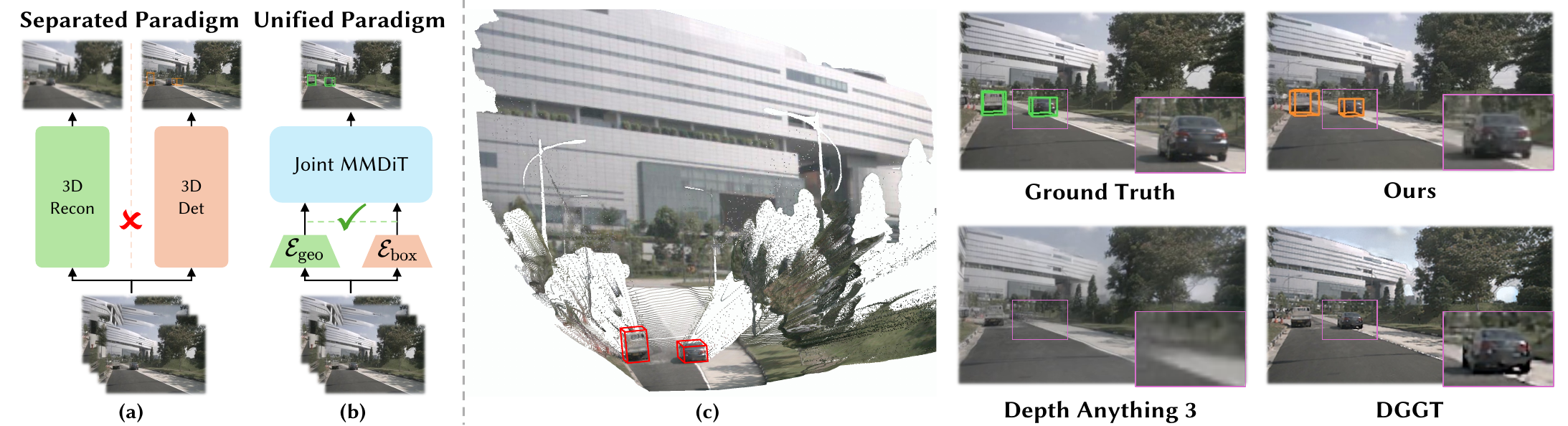}
    \captionof{figure}{\textbf{We present a Unified Generative Framework for 3D Driving Scenes, addressing Reconstruction and Detection as two mutually beneficial representations.} \textbf{(a)} Conventional methods isolate reconstruction from detection, causing temporal smearing and ungrounded boxes. \textbf{(b)} We instead jointly denoise 3D geometry and object-centric layouts as co-evolving latents via a Multi-Modal Diffusion Transformer (MMDiT). \textbf{(c)} Our unified generation mitigates temporal smearing and yields physically consistent, tightly grounded 3D bounding boxes.}
    \label{fig:teaser}
  \end{center}
}]

\makeatletter
\if T\copyright@on
  \insert\aaai@copyrightins{%
    \noindent\footnotesize\copyright@text
  }%
\fi
\makeatother

\begin{abstract}

Spatial representation learning for autonomous driving aims to map raw visual signals into structured 3D scene representations, where object-centric bounding boxes and rendering-oriented 3D primitives (\eg, 3D Gaussians) serve as two distinct yet highly complementary levels for scene understanding. Existing methods typically treat dynamic reconstruction and instance-level perception as separate tasks, despite their shared goal of estimating the underlying 3D world state. As a result, dynamic reconstruction is under-constrained while 3D detection lacks geometric grounding.
To address this gap, we propose \textbf{USR-Drive}, a unified conditional generative framework that, given only posed multi-view driving videos, jointly recovers dense dynamic geometry and instance-level object layouts within a shared scene representation.
Specifically, \textbf{USR-Drive} represents dense Gaussian primitives and sparse 3D bounding boxes as two aligned latent token streams and jointly denoises them with a unified multi-modal diffusion Transformer. Unlike prior paradigms that use boxes as external conditions or predict them with detached modules, \textbf{USR-Drive} treats them as mutually constrained state variables with a Unified Positional Encoding (UPE) that aligns heterogeneous tokens within a shared metric spatiotemporal coordinate.
Via such unified representation and generative framework, the two modalities reinforce each other: geometry supplies dense metric evidence for box prediction, while boxes provide instance-level structural priors that help preserve spatial consistency and reduce ambiguity in sequential 3D geometric representation. Our approach successfully delivers state-of-the-art results for both dynamic reconstruction and 3D detection on the nuScenes and VKitti datasets.

\end{abstract}
\section{Introduction}
\label{sec:intro}

Understanding dynamic 3D scenes from driving video fundamentally requires two tightly coupled capabilities: reconstructing geometrically consistent sequential 3D scene states~\cite{yang2023emernerf, luiten2024dynamic, wang2025vggt, yang2025storm, he2026dynamicvggt, wang2025shape, zhang2025efficiently}, and recovering instance-level 3D detection and tracking~\cite{li2024bevformer, hu2023planning, zhang2022mutr3d}. 
While both areas have advanced rapidly in recent years, they have largely evolved along separate research tracks.

Traditional pipelines treat scene reconstruction and instance-level perception as separate tasks, \eg, for offline rendering and online perception, respectively. However, advances in visual foundation models~\cite{kirillov2023segment,oquab2023dinov2,wang2025vggt,wang2024dust3r,leroy2024grounding} and generative world models~\cite{hu2023gaia,wang2024drivedreamer,li2025end,zuo2025gaussianworld,bogdoll2025muvo} increasingly call for shared representations that support both scene generation and understanding. Despite this trend, driving-scene methods still lack a unified representation that jointly models and mutually constrains dense geometry and object-centric structure.

This gap is particularly limiting for feed-forward reconstruction from driving videos, which is under-constrained by occlusions, large motions, and sparse viewpoints. Without object-level structure, continuous geometry may produce temporally inconsistent geometry and smearing artifacts in dynamic regions (Fig.~\ref{fig:teaser}); conversely, standalone 3D perception lacks explicit grounding in dense reconstructed geometry. Existing unified approaches still model the two asymmetrically: Gaussian-based methods often lack instance-level semantics~\cite{zhu2026worldsplat,mao2025dreamdrive}, while occupancy-based methods use coarse voxel grids that limit rendering fidelity~\cite{bogdoll2025muvo,li2025uniscene}. We therefore represent dense geometry and object layouts as spatially aligned, jointly generated components of a dynamic scene state: geometry provides metric evidence for object localization, while object layouts regularize ambiguous dynamic and occluded regions.

In this paper, we propose USR-Drive, a unified representation for dynamic driving scene reconstruction and instance-level perception from posed multi-view videos. Our key insight is to learn a shared multimodal latent space as a unified spatiotemporal scene state, in which dense, renderable 3D Gaussians and ego-frame 3D bounding boxes are jointly modeled as co-evolving latent variables. USR-Drive first encodes dense spatiotemporal geometry using a pretrained geometric prior~\cite{lin2025da3} and represents 3D bounding boxes as compact object-centric tokens, then maps both streams into a shared metric spatiotemporal coordinate system through our Unified Positional Encoding (UPE), establishing explicit correspondence between dense Gaussian tokens and sparse box tokens. Conditioned on the posed multi-view input video, a unified Multi-Modal Diffusion Transformer (MMDiT) jointly recovers both token streams from Gaussian noise through conditional rectified-flow denoising. This joint process allows dense geometry to provide local metric evidence for object localization, while object-centric tokens supply structural priors that regularize the temporal evolution of reconstructed Gaussians. At inference time, both dynamic 3D Gaussians and 3D boxes are produced directly by the unified denoising process without ground-truth boxes, object tracks, or layout annotations.

Our contributions are summarized as follows:
\begin{itemize}
    \item We propose a unified representation for 3D driving scenes that models object-centric structure and the corresponding dense geometry of the scene as co-evolving state variables, moving beyond detached perception and unconstrained reconstruction paradigms.
    \item We design a dual-branch autoencoding framework to represent heterogeneous scene modalities, coupled with UPE that spatially aligns dense rendering primitives (3D Gaussians) and sparse object-centric structures (3D BBox) within a shared metric spatiotemporal space.
    \item We develop a generative framework with MMDiT that jointly denoises geometry and layout tokens, achieving state-of-the-art performance in both high-fidelity dynamic scene synthesis and temporally coherent 3D detection on various datasets.
\end{itemize}

\section{Related Work}

\subsection{Feed-Forward Scene Reconstruction}
Traditional dynamic scene reconstruction relies on per-scene optimization such as Neural Radiance Fields (NeRFs) or 3D Gaussian Splatting (3DGS), which is computationally intensive and ill-suited for real-time driving applications. Recent work has shifted toward feed-forward architectures that directly infer dense 3D attributes from sparse observations. For static scenes, VGGT~\cite{wang2025vggt} and Depth Anything 3~\cite{lin2025da3} leverage strong visual priors to infer multi-view consistent geometry. For dynamic environments, 4RC~\cite{luo20264rc},D4RT~\cite{zhang2025d4rt} and STORM~\cite{yang2025storm} capture spatiotemporal evolution in a feed-forward manner, while DGGT~\cite{chen2025dggt} achieves pose-free 4D reconstruction with a dynamic head that disentangles moving agents. However, these methods model appearance and geometry purely as low-level rendering primitives. Without explicit instance-level structure such as 3D bounding boxes, the reconstructed dynamic geometry remains weakly constrained and prone to temporal ambiguity and smearing in highly dynamic or occluded traffic scenes.

\subsection{Generative World Models and 3D Perception}
Diffusion models have driven a paradigm shift in 3D perception and scene simulation. Early generative perception methods model 3D bounding boxes directly as denoising trajectories (\eg, 3DifFusionDet~\cite{xiang20263diffusiondet} and MonoDiff~\cite{ranasinghe2024monodiff}), enabling detection without external geometric priors. More recently, world models built on large-scale Diffusion Transformers (DiT)~\cite{wan2025wan} have made layout controllability a central theme: MagicDrive-V2~\cite{gao2025magicdrive} and DrivingDiffusion~\cite{li2024drivingdiffusion} introduce spatio-temporal conditional encodings to guide multi-view video generation with road semantics and 3D boxes. However, these methods use 3D boxes strictly as extrinsic conditioning signals --- the layout serves as an immutable input prior rather than an inferable state variable, and thus cannot be refined from the synthesized geometry.

Pushing toward unified state generation, UniScene~\cite{li2025uniscene} jointly generates 3D semantic occupancy and video, DriveLaW~\cite{xia2025drivelaw} fuses video generation with trajectory planning in a shared latent space, and WorldSplat~\cite{zhu2026worldsplat} predicts feed-forward 4D Gaussians with a 4D-aware latent diffusion model. MagicDrive3D~\cite{gao2024magicdrive3d} couples layout-conditioned video generation with deformable Gaussian reconstruction, while ReconViaGen~\cite{chang2026reconviagen} and Gen3R~\cite{huang2026gen3r} combine feed-forward reconstruction priors with diffusion-based generative priors to improve geometric completeness. Crucially, none of these methods treat sparse 3D bounding boxes and dense continuous geometry as co-predicted, mutually regularized variables within the same denoising process. In contrast, our method co-denoises geometry and layout tokens within a shared world state, establishing bidirectional regularization: dense geometry resolves scale and depth ambiguity for object localization, while object-level structure improves instance-level coherence and rendering rigidity.

\section{Method}
\label{sec:method}
\begin{figure*}[t]
    \centering
    \includegraphics[width=\linewidth]{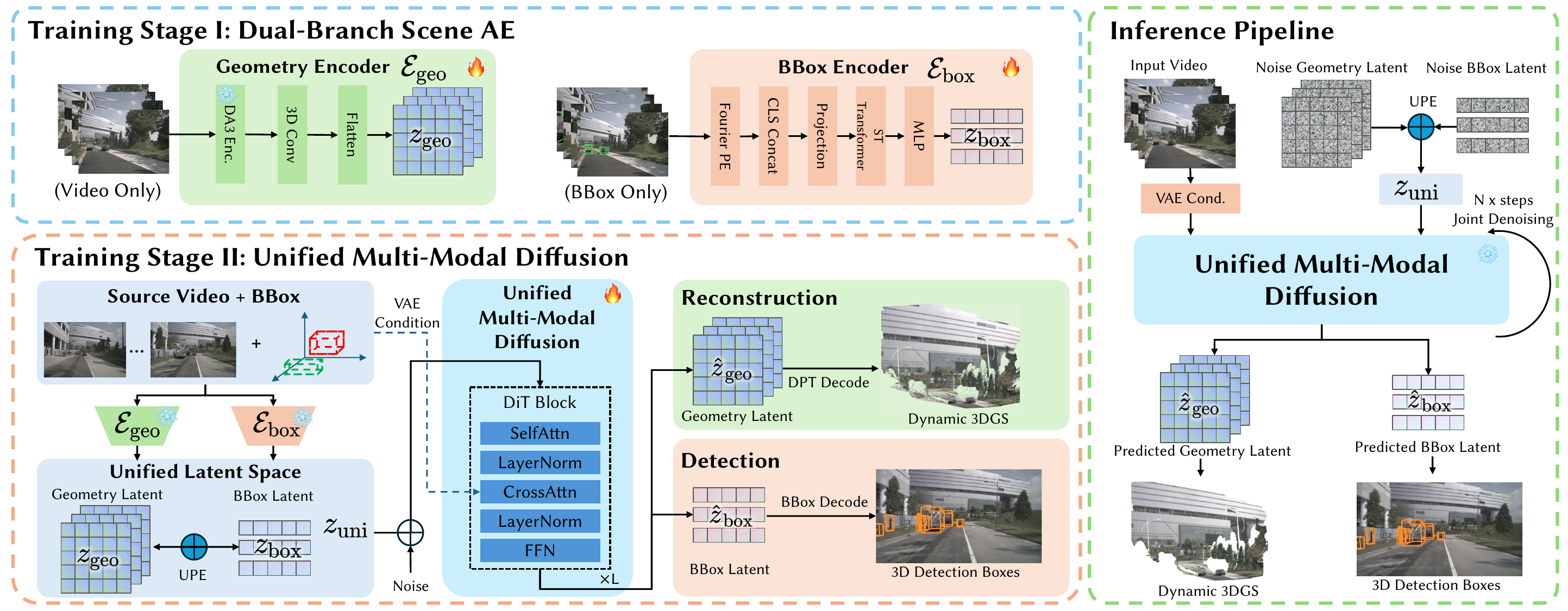}
    \caption{\textbf{Overview of USR-Drive.} \textbf{Stage I (Top):} A dual-branch autoencoder compresses dense geometry and slot-aligned 3D boxes into latents $z_\textrm{geo}$ and $z_\textrm{box}$, respectively. \textbf{Stage II (Bottom):} A unified MMDiT jointly denoises both latent streams, where the Unified Positional Encoding (UPE) anchors heterogeneous tokens in a common 3D metric space, enabling bidirectional regularization between geometry and layout through shared self-attention. \textbf{Inference (Right):} Starting from pure noise, the MMDiT iteratively denoises the joint latent streams conditioned on reference video latents, and the denoised latents are decoded by the DPT geometry head and the box decoder into explicit dense scene geometry and 3D object layouts.}
    \label{fig:overview}
\end{figure*}

\subsection{Overview}
\label{sec:overview}
USR-Drive aims to unify dense geometric recovery and instance-level 3D object detection within a joint multi-modal framework. As illustrated in Fig.~\ref{fig:overview}, the overall framework consists of two tightly coupled stages. In the first stage, we learn a \textit{dual-branch autoencoder} system that compresses two heterogeneous yet complementary modalities into compact latent representations: a geometric branch that captures dense scene structure and temporal dynamics from the raw video, and a layout branch that encodes object-level spatial information from bounding box sequences. These two branches yield a disentangled yet aligned latent space prepared for downstream co-generation.

In the second stage, we build a \textit{unified multi-modal diffusion architecture} over the learned latent spaces. Instead of performing diffusion in a low-dimensional, patch-aligned latent space, our model operates on the joint latent tokens from both the geometry and layout branches, which preserve richer scene information. The diffusion model is trained to perform joint denoising for scene-level geometric reconstruction and object-level layout grounding.

\subsection{Dual-branch Scene Autoencoder}

Dynamic driving scenes involve heterogeneous representations ranging from dense geometry to sparse object-centric structures. To model these complementary modalities, we adopt a dual-branch autoencoder framework that separately encodes dense Gaussian geometry and 3D bounding box layouts before aligning them in a unified metric spatiotemporal latent space for joint diffusion modeling.

\paragraph{\textbf{Dense Geometry Autoencoder}}
Instead of learning a latent space directly from raw 3D Gaussian parameters, our geometry encoder $\mathcal{E}_\textrm{geo}$ leverages a frozen, pretrained foundational geometry encoder, Depth Anything V3 Base (DA3-Base)~\cite{lin2025da3}, to extract high-fidelity geometric features as the continuous dense geometry representation $z_\textrm{geo}^c$.
To ensure the resulting feature space forms a well-regularized and smooth manifold suitable for latent diffusion, we formulate this bottleneck as a Representation Autoencoder (RAE)~\cite{zheng2026rae}, which projects $z_\textrm{geo}^c$ into the MMDiT hidden space through a lightweight 3D convolutional encoder, yielding the geometry latent $z_\textrm{geo} \in \mathbb{R}^{B \times T \times N_\textrm{geo} \times D_\textrm{g}}$.
For decoding, we symmetrically employ the Dense Prediction Transformer (DPT) decoder~\cite{ranftl2021dpt} from DA3 to reconstruct the dense scene geometry from the latents.
The geometry autoencoder is trained end-to-end with a composite objective that supervises both photometric rendering fidelity ($\mathcal{L}_\textrm{rgb}$) and geometric structure via sparse depth regularization from ground-truth LiDAR maps ($\mathcal{L}_\textrm{depth}$). Detailed layer configurations and loss formulations are provided in the \emph{supplementary materials}.

\paragraph{\textbf{Sparse Bounding Box Autoencoder}}
Parallel to the dense geometry branch, discrete object-level constraints are modeled via a bounding box path. Our BBox Autoencoder $\mathcal{E}_\textrm{box}$ encodes slot-aligned 3D bounding box sequences into latents $z_\textrm{box} \in \mathbb{R}^{B \times T \times N_{\mathrm{max}} \times D_\textrm{b}}$, where $N_{\mathrm{max}}$ is the maximum number of boxes. Specifically, each box is converted into its eight 3D corners, embedded with a Fourier Positional Embedding (FPE) and a learnable class embedding, and processed by spatiotemporal transformer blocks that alternately perform frame-local spatial attention across object slots and temporal attention along each slot's trajectory. A symmetric decoder then reconstructs the explicit 3D bounding box parameters from $z_\textrm{box}$.

The BBox Autoencoder is trained independently following established 3D layout representation paradigms~\cite{gao2025magicdrive}, where invalid slots are handled by a binary mask. The reconstruction objective combines an $\ell_1$ corner regression loss, a sine-cosine yaw regression loss in the $\mathrm{SO}(2)$ embedding space to circumvent periodicity artifacts, a cross-entropy category loss, a slot-existence loss, and a KL regularizer over the latent distribution. Please refer to the \emph{supplementary materials} for detailed formulations and training strategies.

\subsection{Unified Diffusion Architecture}

\paragraph{\textbf{Unified Positional Encoding}}

\begin{figure}[t]
    \centering
    \includegraphics[width=\linewidth]{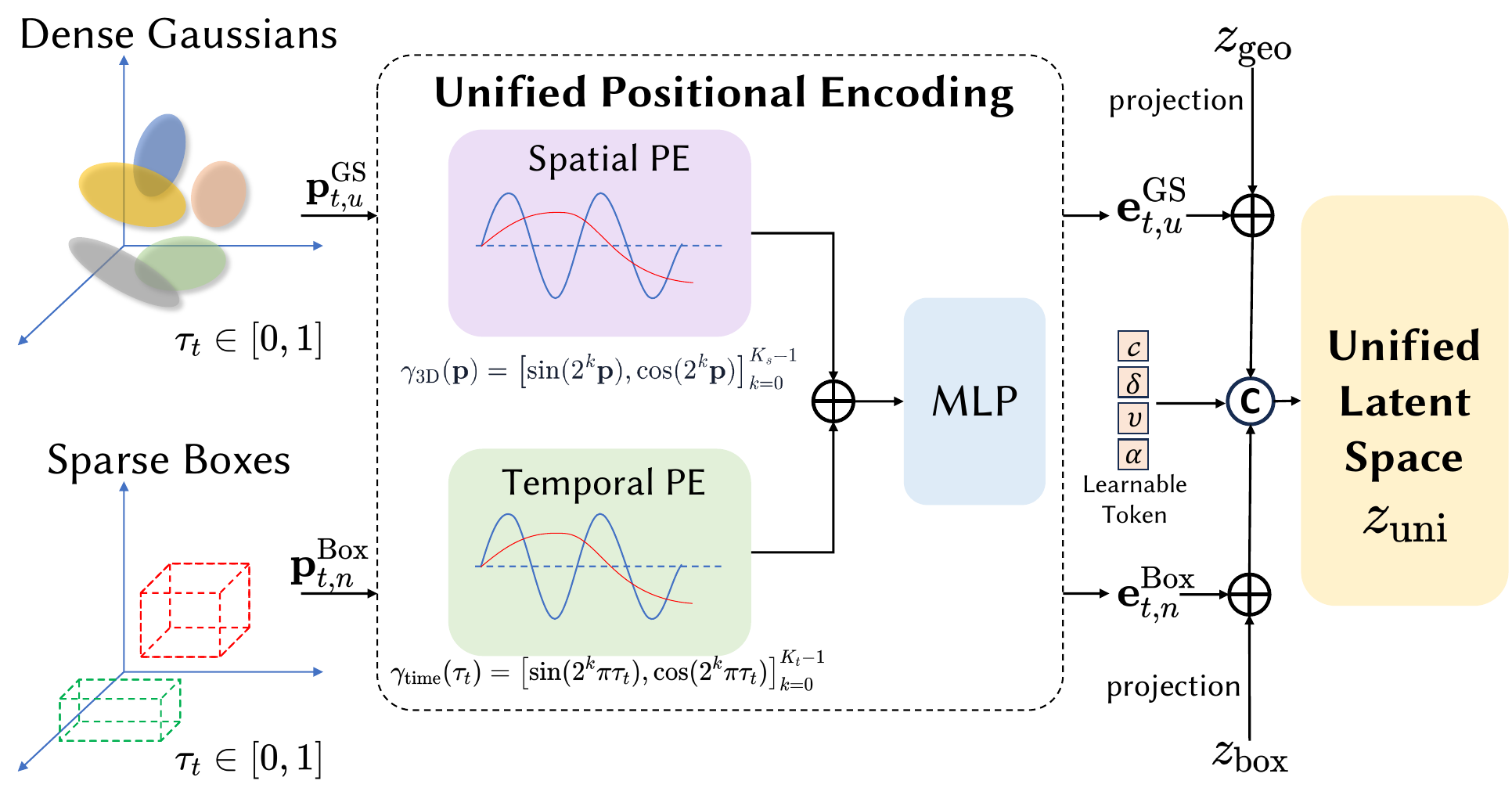}
    \caption{\textbf{Unified Positional Encoding.} Dense GS tokens $\mathbf{p}^{\mathrm{GS}}_{t,u}$ and sparse box tokens $\mathbf{p}^{\mathrm{Box}}_{t,n}$ are assigned metric 3D anchors and a normalized frame index, which are encoded by shared spatial and temporal Fourier features followed by an MLP. The resulting UPE embeddings are combined with projected geometry and box latents together with learnable modality-aware tokens, aligning heterogeneous modalities into a unified latent space for joint DiT modeling.}
    \label{fig:upe}
\end{figure}

In the second stage, we perform denoising process over two heterogeneous latent streams: dense Gaussian Splatting (GS) geometry tokens and sparse 3D bounding-box layout tokens. Since the two modalities differ in token density, semantics, and ordering, independent index-based positional embeddings cannot provide meaningful cross-modal spatial correspondence. As shown in Fig.~\ref{fig:upe}, we introduce a Unified Positional Encoding (UPE), which represents both modalities in a shared metric spatio-temporal scene coordinate system.

Given a token with metric 3D anchor $\mathbf{p}\in\mathbb{R}^{3}$ at frame $t$, UPE encodes its spatial location and temporal index jointly through
Fourier features:
\begin{equation}
\begin{aligned}
\Phi_{\mathrm{upe}}(\mathbf{p},t)
&=\mathrm{MLP}_{\mathrm{upe}}
\left(\left[\gamma_{\mathrm{3D}}(\mathbf{p}),
\gamma_{\mathrm{time}}(\tau_t)\right]\right),\\
\tau_t&=\frac{t}{T-1}.
\end{aligned}
\end{equation}

Here, $\tau_t\in[0,1]$ denotes the normalized frame index, independent of the diffusion timestep. The spatial and temporal Fourier features are defined as:
\begin{equation}
\begin{aligned}
\gamma_{\mathrm{3D}}(\mathbf{p})
& =
\left[
\sin(2^k\mathbf{p}),
\cos(2^k\mathbf{p})
\right]_{k=0}^{K_s-1},
\\
\gamma_{\mathrm{time}}(\tau_t)
& =
\left[
\sin(2^k\pi\tau_t),
\cos(2^k\pi\tau_t)
\right]_{k=0}^{K_t-1},
\end{aligned}
\end{equation}
where $K_s$ and $K_t$ denote the number of Fourier frequency bands used for spatial coordinates and temporal indices, respectively.
This shared encoding maps both geometry and layout tokens into a unified geometry-centric positional space.


Specifically, during the training phase, we construct metric anchors for GS tokens using only the reference video, its calibrated camera poses, and a frozen geometry prior.
Before denoising, the geometry prior is evaluated once to obtain a coarse set of Gaussian primitives, including their preliminary centers $\widetilde{\boldsymbol{\mu}}^{\mathrm{prior}}_{t,q}$ and opacities $\widetilde{\alpha}^{\mathrm{prior}}_{t,q}$.
Using the known camera poses, we transform each preliminary center into the ego coordinate system of the first frame and denote the transformed center by $\overline{\boldsymbol{\mu}}^{\mathrm{prior}}_{t,q}$. For each GS token $u$, we then define its anchor as the opacity-weighted centroid of the prior primitives associated with its corresponding image-space patch $\mathcal{P}(u)$. 
Importantly, this anchor provides only a coarse metric reference for the token. It should not be interpreted as either the clean diffusion target or the Gaussian center predicted by the diffusion model.

For box tokens, instead of using ground-truth box centers during training, we assign each slot to a deterministic BEV grid anchor:
\begin{equation}
\mathbf{p}^{\mathrm{GS}}_{t,u}
=
\frac{
\sum_{q\in\mathcal{P}(u)}
\widetilde{\alpha}^{\mathrm{prior}}_{t,q}
\overline{\boldsymbol{\mu}}^{\mathrm{prior}}_{t,q}
}{
\sum_{q\in\mathcal{P}(u)}
\widetilde{\alpha}^{\mathrm{prior}}_{t,q}
+\epsilon
},
\qquad
\mathbf{e}^{\mathrm{GS}}_{t,u}
=
\Phi_{\mathrm{upe}}
\left(
\mathbf{p}^{\mathrm{GS}}_{t,u},t
\right),
\end{equation}
\vspace{-2mm}
\begin{equation}
\mathbf{p}^{\mathrm{Box}}_{t,n}
=
\mathbf{p}^{\mathrm{BEV}}_{n},
\qquad
\mathbf{e}^{\mathrm{Box}}_{t,n}
=
\Phi_{\mathrm{upe}}
\left(
\mathbf{p}^{\mathrm{Box}}_{t,n},t
\right).
\end{equation}
where $\overline{\boldsymbol{\mu}}^{\mathrm{prior}}_{t,q}$ denotes the preliminary center of Gaussian $q$ after transformation into the $t=0$ ego coordinate system, and $\widetilde{\alpha}^{\mathrm{prior}}_{t,q}$ denotes its opacity.
$\mathcal{P}(u)$ denotes the set of prior Gaussian primitives associated with geometry token $u$. The geometry anchors and the fixed BEV anchors are therefore expressed in the same metric reference coordinate system.
$\mathbf{p}^{\mathrm{BEV}}_{n}$ denotes the metric center of the $n$-th cell in a fixed BEV grid and is shared across all frames; its temporal instances are distinguished by the frame index $t$ in $\Phi_{\mathrm{upe}}(\mathbf{p},t)$.

Neither type of anchor depends on the clean diffusion targets. The GS anchors are computed once during preprocessing, whereas the BEV anchors are predefined; both remain fixed throughout the denoising process.

Furthermore, to improve detection accuracy and motion perception, we augment each box latent with a confidence scalar $c$, an anchor-relative position bias $\delta$, a velocity vector $v$, and an attribute vector $a$ for dynamic judgment. We concatenate $[z_\textrm{uni}, c, \delta, v, a]$ as the augmented unified latent and send it to the DiT, which predicts the corresponding flows jointly with the geometry and box latents. During decoding, the predicted $\hat{\delta}$ is mapped back to metric space to refine the final box center, while $\hat{v}$ and $\hat{a}$ serve as the box velocity and dynamic status predictions.
Please refer to our \emph{supplementary materials} for detailed designs and ablations.


\paragraph{\textbf{Multi-Modal Diffusion Transformer}}
Inspired by recent geometry RAE based diffusion models~\cite{jang2026gld}, we design a unified multi-modal diffusion model to directly denoise the joint noisy latent space. Specifically, we adopt the structure of DiT blocks from Wan 2.1-1.3B~\cite{wan2025wan}. The spatially grounded geometry and bounding box tokens are concatenated into a singular 1D sequence and processed through a unified Multi-Modal Diffusion Transformer (MMDiT) attention trunk. To guide the denoising process, the network is conditioned on reference video latents extracted from a visual VAE, which are linearly projected and injected as the cross-attention context. The denoised latents are subsequently processed by their respective autoencoder decode heads --- the DA3-based DPT head and the symmetric bounding box decoder --- to yield the final explicit 3D geometry and discrete object layouts.

\paragraph{\textbf{Training Strategy}}
We formulate the generative training objective within the continuous-time rectified flow framework~\cite{lipman2022flow}, which predicts the target flow velocity pointing from the data sample toward the noise sample. 

Given the predicted velocities from the geometry and layout heads, denoted by $v_{\mathrm{geo}}$ and $v_{\mathrm{box}}$, respectively, we optimize the two branches with separate objectives to account for their different structures. For the geometry branch, we apply a standard Mean Squared Error (MSE) loss $\mathcal{L}_{\mathrm{geo}} = \mathrm{MSE}(\mathbf{v}_\textrm{geo}, \tilde{\mathbf{v}}_\textrm{geo})$ between the predicted and target velocities.

For the layout branch, supervision is applied only to valid bounding box slots. Let $M \in \{0,1\}^{N_{\max}}$ denote the binary occupancy mask for the box tokens, where $M_i=1$ indicates a valid object slot. The layout loss is defined as:
\begin{equation}
\mathcal{L}_{\mathrm{box}}
=
\frac{
\sum_{i=1}^{N_{\max}}
M_i \,
\left\|
\mathbf{v}_\textrm{box}^i - \tilde{\mathbf{v}}_\textrm{box}^i
\right\|_2^2
}{
\sum_{i=1}^{N_{\max}} M_i
},
\end{equation}
where $\tilde{\mathbf{v}}_\textrm{box}^i$ is the target velocity for the $i$-th box slot. This masking ensures that the layout objective is evaluated only on populated entities, which is particularly important for stable training under slot masking augmentation.

In addition to the latent layout velocity, we supervise auxiliary layout targets on the fixed-anchor slots, including object confidence, anchor-relative center offset, ground-plane velocity, and object attribute. We group these terms into an auxiliary layout loss:
\begin{equation}
\mathcal{L}_{\mathrm{aux}}
=
\lambda_{{c}}\mathcal{L}_{{c}}
+
\lambda_{\delta}\mathcal{L}_{\delta}
+
\lambda_{{v}}\mathcal{L}_{{v}}
+
\lambda_{{a}}\mathcal{L}_{{a}} .
\end{equation}

The overall training objective is:
\begin{equation}
\mathcal{L}_{\mathrm{total}}
=
\mathcal{L}_{\mathrm{geo}}
+
\lambda_{\mathrm{box}}\mathcal{L}_{\mathrm{box}}
+
\mathcal{L}_{\mathrm{aux}} .
\label{total_loss}
\end{equation}
Please refer to our \emph{supplementary materials} for detailed loss function.
%

\section{Experiments}
\label{sec:exp}

\begin{figure*}[t]
    \centering
    \includegraphics[width=0.94\linewidth]{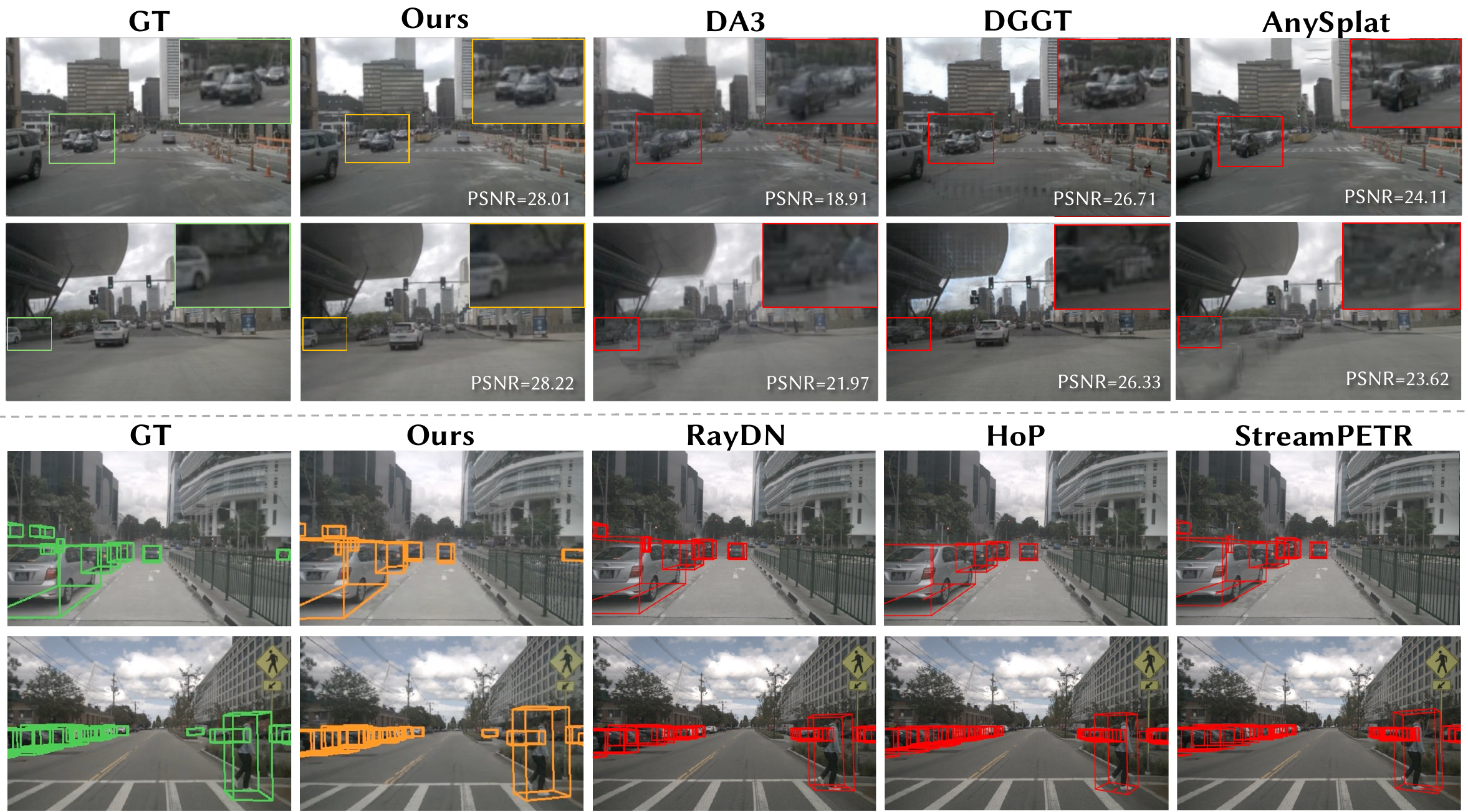}
    \caption{\textbf{Visualization of reconstruction and detection results.} Compared to baseline methods, our approach reveals better quality on both reconstruction task (top 2 rows) and detection task (bottom 2 rows).}
    \label{fig:main_exp}
\end{figure*}

\subsection{Experimental Setup}
\paragraph{\textbf{Dataset}}
We conduct experiments on the nuScenes~\cite{caesar2020nuscenes} and VKitti~\cite{gaidon2016virtual} datasets, which provide synchronized camera streams and 3D annotations for autonomous driving.
We use the nuScenes official split for training and evaluation. We take all $6$ surround-view cameras as input with temporal clips of $T{=}8$ frames, and all images are resized to $112\times168$; camera poses are used only for metric grounding of the UPE anchors.
As for VKitti dataset, we sampled 400 cases for zero-shot evaluation.
Please refer to \emph{supplementary materials} for detailed dataset preprocessing. 

\paragraph{\textbf{Training and Inference}}
All models are trained on the nuScenes dataset solely.
During inference, the only scene-content input is the posed multi-view video: no ground-truth boxes, object tracks, or layout annotations are used. We encode the video with the frozen Wan-VAE as the conditioning signal and initialize all latents ($z_\textrm{geo}, z_\textrm{box}, c, \delta, v, a$) from Gaussian noise, denoising them with $50$ rectified-flow steps under classifier-free guidance between the video-conditioned and zero-conditioned predictions. The sparse layout tokens are tied to a deterministic BEV anchor grid rather than ground-truth box centers, so boxes are pure outputs of the unified denoising process. The denoised geometry and layout latents are then decoded by the DPT decoder and the frozen BBox-AE into 3D Gaussians and 3D bounding boxes, respectively; longer sequences are handled with overlapping $8$-frame sliding windows. Detailed decoding and post-processing procedures are provided in the \emph{supplementary materials}.


\paragraph{\textbf{Baselines}}
We evaluate our method on two complementary tasks: 
(1) dynamic scene reconstruction, and 
(2) 3D object detection. 
This dual evaluation directly reflects our goal of unifying geometry and perception. 

We compare against two categories of methods:

(1) \textit{Reconstruction methods.}
We compare our method with the recent geometry foundation models including VGGT~\cite{wang2025vggt}, Depth Anything 3~\cite{lin2025da3} and Pi-3~\cite{wang2026pi3}. In addition, we also compare with the state-of-the-art feed-forward Gaussian reconstruction methods including AnySplat~\cite{jiang2025anysplat}, STORM~\cite{yang2025storm} and DGGT~\cite{chen2025dggt}.

(2) \textit{3D detection methods.}
We report the 3D detection results of nuScenes on their val set, including \textbf{the best performed} methods reported in the official benchmark (\eg, HoP~\cite{zong2023hop}, StreamPETR~\cite{wang2023exploring}, RayDN~\cite{liu2024ray}, \emph{etc.}). As for VKitti dataset, we calculate mAP and mATE metrics with the ground truth 3D annotations.


The detailed implementations of the geometry and bounding box autoencoders and the unified diffusion model are provided in the \emph{supplementary materials}.

\subsection{Experimental Results}

\paragraph{\textbf{Reconstruction Quality}}
We first evaluate dynamic scene reconstruction quality against recent geometry foundation models and dynamic Gaussian reconstruction methods. As reported in Tab.~\ref{tab:recon_gen}, our method achieves the best performance across all image-level metrics and 3D depth metrics. In particular, the gain is most pronounced on foreground dynamic objects, where reconstruction is most under-constrained: USR-Drive improves object-level PSNR and SSIM by a clear margin over all baselines. Since dynamic regions benefit the least from appearance cues alone, this improvement directly evidences that the sparse box tokens serve as instance-level structural priors that regularize dynamic geometry --- a concrete source of the mutual benefit enabled by joint denoising. We attribute this improvement to the unified latent modeling of dense geometry and sparse object layout, which helps preserve scene structure while reducing local visual artifacts. Qualitative results on nuScenes further show that our method recovers better scene details, cleaner object boundaries, and more coherent dynamic content under challenging driving scenarios. Please refer to Fig. \ref{fig:main_exp} and \emph{supplementary materials} for visualization results.

\begin{table}[t]
    \centering
    \small
    \setlength{\tabcolsep}{4pt}
    
    \begin{tabular}{l|ccc|c}
        \toprule
        \textbf{Method}
        & \textbf{PSNR}$\uparrow$
        & \textbf{SSIM}$\uparrow$
        & \textbf{LPIPS}$\downarrow$
        & \textbf{D-RMSE}$\downarrow$ \\
        \midrule

        \multicolumn{5}{c}{
            \textit{Scene-level Reconstruction}
        } \\
        \midrule
        VGGT
        & -- & -- & --
        & \cellcolor{yellow!30}5.07 \\

        DA3
        & 21.75 & 0.617 & 0.183 & 12.16 \\

        Pi-3
        & -- & -- & -- & 6.93 \\

        AnySplat
        & \cellcolor{orange!30}25.53
        & \cellcolor{orange!30}0.803
        & \cellcolor{orange!30}0.157
        & 17.65 \\

        STORM
        & 24.54 & 0.784 & 0.267 & 6.48 \\

        DGGT
        & \cellcolor{yellow!30}26.63
        & \cellcolor{yellow!30}0.813
        & \cellcolor{yellow!30}0.122
        & \cellcolor{orange!30}5.08 \\

        \textbf{USR-Drive}
        & \cellcolor{green!30}\textbf{27.55}
        & \cellcolor{green!30}\textbf{0.853}
        & \cellcolor{green!30}\textbf{0.076}
        & \cellcolor{green!30}\textbf{4.59} \\

        \midrule
        \multicolumn{5}{c}{
            \textit{Object-level Foreground Reconstruction}
        } \\
        \midrule

        AnySplat
        & 14.78 & 0.395 & \cellcolor{orange!30}0.209 & 19.93 \\

        DA3
        & 14.34 & 0.408 & 0.297 & \cellcolor{orange!30}15.32 \\

        STORM
        & \cellcolor{yellow!30}20.97
        & \cellcolor{orange!30}0.532
        & 0.515 & \cellcolor{yellow!30}12.56 \\

        DGGT
        & \cellcolor{orange!30}19.73
        & \cellcolor{yellow!30}0.791
        & \cellcolor{yellow!30}0.150 & 19.78 \\

        \textbf{USR-Drive}
        & \cellcolor{green!30}\textbf{24.45}
        & \cellcolor{green!30}\textbf{0.833}
        & \cellcolor{green!30}\textbf{0.083} & \cellcolor{green!30}\textbf{9.98} \\

        \bottomrule
    \end{tabular}
    \caption{
        \textbf{Scene-level and object-level reconstruction results on the
        nuScenes dataset.}
        The object-level evaluation is conducted on foreground dynamic
        objects. We highlight the
        \colorbox{green!30}{\textbf{first}},
        \colorbox{yellow!30}{second}, and
        \colorbox{orange!30}{third} {best results within each reconstruction block.}
        VGGT and Pi-3 only produce 3D point clouds, so rendered-image
        quality metrics are omitted.
    }
    \label{tab:recon_gen}
\end{table}

\paragraph{\textbf{3D Detection Quality}}
We follow the official nuScenes detection benchmark protocol for the vision track. The results on the nuScenes benchmark reveal the high precision and recall of our method for 3D detection. As shown in Tab.~\ref{tab:detection_val}, our method surpasses dedicated state-of-the-art detectors on both NDS and mAP. Visualization results in Fig.\ref{fig:main_exp} and the \emph{supplementary materials} further demonstrate that our joint generation paradigm benefits from the information of the geometric branch, achieving more accurate and complete object-level detection. Moreover, the zero-shot detection results on the VKitti datasets in Tab.~\ref{tab:vkitti_unified_results} also demonstrate the strong generalizability of our method, while baseline methods fail to generalize to the unseen synthetic domain.

\begin{table}[t]
\centering
\small
\setlength{\tabcolsep}{1pt}
\begin{tabular}{l|ccccccc}
\toprule
\textbf{Method} & \scriptsize{\textbf{NDS}$\uparrow$} & \scriptsize{\textbf{mAP}$\uparrow$} & \scriptsize{\textbf{mATE}$\downarrow$} & \scriptsize{\textbf{mASE}$\downarrow$} & \scriptsize{\textbf{mAOE}$\downarrow$} & \scriptsize{\textbf{mAVE}$\downarrow$} & \scriptsize{\textbf{mAAE}$\downarrow$} \\
\midrule
\scriptsize{BEVDepth}  & 0.475 & 0.351 & 0.629 & 0.267 & 0.479 & 0.428 & 0.198 \\
\scriptsize{BEVFormer}  & 0.517 & 0.416 & 0.673 & 0.274 & 0.372 & 0.394 & 0.198 \\
\scriptsize{PETRv2} & 0.456 & 0.349 & 0.700 & 0.275 & 0.580 & 0.437 & \cellcolor{yellow!30}0.187 \\
\scriptsize{HoP} & 0.558 & 0.454 & \cellcolor{orange!30}0.565 & 0.265 & \cellcolor{orange!30}0.327 & 0.337 & \cellcolor{orange!30}0.194 \\
\scriptsize{StreamPETR} & 0.550 & 0.450 & 0.613 & 0.267 & 0.413 & 0.265 & 0.198 \\
\scriptsize{RayDN} & \cellcolor{orange!30}0.563 & \cellcolor{orange!30}0.469 & 0.579 & \cellcolor{orange!30}0.264 & 0.433 & \cellcolor{orange!30}0.256 & \cellcolor{yellow!30}0.187 \\
\scriptsize{RoPETR}  & \cellcolor{yellow!30}0.614 & \cellcolor{yellow!30}0.529 & \cellcolor{yellow!30}0.537 & \cellcolor{yellow!30}0.255 & \cellcolor{green!30}\textbf{0.289} & \cellcolor{green!30}\textbf{0.229} & 0.195 \\
\midrule
\scriptsize{\textbf{USR-Drive}}  & \cellcolor{green!30}\textbf{0.625}  & \cellcolor{green!30}\textbf{0.552}  & \cellcolor{green!30}\textbf{0.525}  & \cellcolor{green!30}\textbf{0.211}  & \cellcolor{yellow!30}0.303 & \cellcolor{yellow!30}0.251 & \cellcolor{green!30}\textbf{0.177} \\
\bottomrule
\end{tabular}
\caption{\textbf{Results of 3D detection tasks on the nuScenes val set.} We inherit the reported results of baseline methods from their official publications~\cite{zong2023hop, liu2024ray, ji2025ropetr}.}
\label{tab:detection_val}
\end{table}

\begin{table}[t]
    \centering
    \small
    \setlength{\tabcolsep}{3.5pt}
        \begin{tabular}{l|cc|cc}
        \toprule
        \multirow{2}{*}{\textbf{Method}} & \multicolumn{2}{c|}{\textbf{Reconstruction}} & \multicolumn{2}{c}{\textbf{Detection}} \\
        \cmidrule{2-5}
         & \textbf{PSNR$\uparrow$} & \textbf{SSIM$\uparrow$} & \textbf{mAP$\uparrow$} & \textbf{mATE$\downarrow$} \\
        \midrule
        \multicolumn{5}{c}{\textit{Detection-Only Baselines}} \\
        \midrule
        RayDN & - & - & \cellcolor{orange!30}0.015 & \cellcolor{yellow!30}1.024 \\
        HoP & - & - & \cellcolor{yellow!30}0.022 & \cellcolor{orange!30}1.258 \\
        StreamPETR & - & - & 0.008 & 1.340 \\
        \midrule
        \multicolumn{5}{c}{\textit{Reconstruction-Only Baselines}} \\
        \midrule
        DA3 & 17.12 & 0.465 & - & - \\
        DGGT & \cellcolor{yellow!30}25.38 & \cellcolor{yellow!30}0.712 & - & - \\
        AnySplat & \cellcolor{orange!30}23.80 & \cellcolor{orange!30}0.625 & - & - \\
        \midrule
        \textbf{USR-Drive}
        & \cellcolor{green!30}\textbf{26.45}
        & \cellcolor{green!30}\textbf{0.743}
        & \cellcolor{green!30}\textbf{0.518}
        & \cellcolor{green!30}\textbf{0.812} \\
        \bottomrule
    \end{tabular}
    \caption{\textbf{Zero-shot reconstruction and 3D detection results on VKitti.} Our unified model performs both tasks, while baselines specialize in only one.}
\label{tab:vkitti_unified_results}
\end{table}

\subsection{Ablation Studies}
\paragraph{\textbf{Joint Denoising vs.\ Decoupled Pipelines}}
To verify that our gains stem from the unified co-denoising design rather than the pretrained DA3 backbone alone, we compare against (i) the Stage-I geometry autoencoder without diffusion refinement, (ii) a decoupled two-stage pipeline that first reconstructs geometry and then predicts boxes from it, and (iii) a cascaded DA3 + StreamPETR~\cite{wang2023exploring} pipeline for detection. As shown in Tab.~\ref{tab:ablation_upe}, the decoupled variants degrade both tasks: errors in the two-stage pipeline compound across stages, and the box branch can no longer feed structural priors back into the geometry. In contrast, UPE places noisy geometry and box tokens in a shared metric space, where MMDiT self-attention allows them to exchange dense geometric evidence and object-level layout cues during denoising, yielding mutual correction.

\paragraph{\textbf{Component and UPE Ablation}}
We further ablate the key components of our framework, as reported in the bottom block of Tab.~\ref{tab:ablation_upe}. Removing the layout branch, which reconstructs scenes using only finetuned DA3 geometry features(Geom. Only), degrades all reconstruction metrics, demonstrating that sparse layout tokens provide complementary structural cues that help resolve scene ambiguities during joint denoising. Conversely, relying exclusively on the discrete bounding box branch collapses 3D detection to a mere $0.012$ mAP: without dense geometric features providing structural context and spatial grounding, predicting object layouts in isolation is highly ill-posed. Finally, removing the metric spatio-temporal positional encoding leaves heterogeneous tokens coordinated only by sequence order, causing clear degradation in both layout accuracy and reconstruction quality. These results confirm that the two modalities mutually reinforce each other and that UPE provides a crucial shared spatial reference for aligning them. Visualized ablation results are provided in the \emph{supplementary materials}.

\begin{table}[t]
    \centering
    \setlength{\tabcolsep}{1.75pt}
    \small
    \begin{tabular}{l|ccc|cc}
        \toprule
        \textbf{Method}
        & {\scriptsize\textbf{PSNR}$\uparrow$}
        & {\scriptsize\textbf{SSIM}$\uparrow$}
        & {\scriptsize\textbf{LPIPS}$\downarrow$}
        & {\scriptsize\textbf{mAP}$\uparrow$}
        & {\scriptsize\textbf{mATE}$\downarrow$} \\
        \midrule

        \multicolumn{6}{c}{
            \textit{Joint Denoising and Pipeline Comparison}
        } \\
        \midrule
        Stage-I Geo-AE
        & 23.43
        & 0.709
        & 0.177
        & -- & -- \\

        Decoupled two-stage
        & 21.47
        & 0.655
        & 0.179
        & 0.473
        & 0.701 \\

        DA3 + StreamPETR
        & -- & -- & --
        & 0.491
        & 0.648 \\


        \midrule
        \multicolumn{6}{c}{
            \textit{Component and UPE Ablation}
        } \\
        \midrule
        Geom. Only
        & 26.37
        & 0.833
        & 0.127
        & -- & -- \\

        Box Only
        & -- & -- & --
        & 0.012
        & 0.901 \\

        Ours w/o UPE
        & 25.32
        & 0.799
        & 0.134
        & 0.214
        & 0.602 \\
        \midrule
        \textbf{Ours}
        & \textbf{27.55}
        & \textbf{0.853}
        & \textbf{0.076}
        & \textbf{0.552}
        & \textbf{0.525} \\

        \bottomrule
    \end{tabular}
    \caption{
        \textbf{Ablation results on nuScenes val set.}
        Top: joint denoising vs.\ decoupled pipelines.
        Bottom: component and UPE ablations.
        Our full model performs best on both tasks.
        }
    \label{tab:ablation_upe}
\end{table}


\paragraph{\textbf{More Ablation Studies}}
We also conduct ablation studies on the \textbf{auxiliary layout tokens}, \textbf{training parameters}, and \textbf{MMDiT hidden dimension}. Please refer to our \textit{supplementary materials} for detailed discussion.

\section{Conclusion}
\label{sec:conclusion}
In this paper, we propose USR-Drive, a unified generative framework for autonomous driving that jointly models dynamic 3D geometry and instance-level 3D bounding boxes as a shared physical world state. By introducing a multi-modal latent space, a Unified Positional Encoding (UPE), and a joint Diffusion Transformer (DiT) for simultaneous denoising, our method enables dense geometry and object structure to mutually constrain each other. This unified design improves both the fidelity and appearance consistency of dynamic reconstruction and the geometric grounding of 3D detection. Experiments on nuScenes and VKitti demonstrate state-of-the-art performance on both tasks, highlighting the promise of unified scene representations for future autonomous driving world models.

\paragraph{\textbf{Limitations and Future Work}}
Our patch- and frame-aligned geometry representation captures local consistency but lacks a compact global scene state and explicit long-term identities, limiting dynamic modeling and direct support for 4D tracking. Future work will explore global representations that unify reconstruction, detection, and tracking. Moreover, iterative denoising introduces non-trivial latency, restricting the current model to offline estimation. Few-step distillation or causal autoregressive generation may enable real-time deployment and long-term scene representation.

\bibliography{main}

\clearpage
\appendix
\section{Dual-branch Scene Autoencoder Details}
\label{sec:supp_autoencoder}

\subsection{Dense Geometry Autoencoder}
The geometry encoder $\mathcal{E}_\textrm{geo}$ builds on the frozen Depth Anything V3 Base (DA3-Base) encoder~\cite{lin2025da3}. To reduce the token space and alleviate GPU memory consumption, we extract the token representations from the $5^\textrm{th}$ layer of the DA3-Base ViT backbone as the continuous dense geometry representation $z_\textrm{geo}^c$.
On top of these features, the RAE bottleneck applies a lightweight 3D convolutional encoder with a $1 \times 1 \times 1$ kernel, which projects each spatio-temporal feature location into the MMDiT hidden space, producing the final geometry latent $z_\textrm{geo} \in \mathbb{R}^{B \times T \times N_\textrm{geo} \times D_\textrm{g}}$, where $N_\textrm{geo}=H_\textrm{geo} \times W_\textrm{geo}$. Unlike standard VAEs that often induce high-frequency detail loss via strict KL-divergence regularization, the RAE formulation maintains a well-regularized and smooth feature manifold.
The geometry autoencoder is trained end-to-end with a composite objective:
\begin{equation}
\mathcal{L}_\textrm{Geo-AE} = \lambda_\textrm{rgb} \mathcal{L}_\textrm{rgb} + \lambda_\textrm{depth} \mathcal{L}_\textrm{depth},
\end{equation}
where $\mathcal{L}_\textrm{rgb}$ applies an $\mathcal{L}_1$ penalty between the rendered and ground-truth RGB images, and $\mathcal{L}_\textrm{depth}$ provides sparse depth regularization from ground-truth LiDAR maps to maintain structural accuracy.
The geometry autoencoder is fine-tuned from the pretrained DA3 weights, and temporal interactions among the flattened geometry tokens are modeled by the subsequent MMDiT attention blocks.

\subsection{Sparse Bounding Box Autoencoder}
The BBox Autoencoder is trained independently before joint diffusion training. For each frame, object tracks are padded or truncated to a fixed number of slots, where each valid slot contains a 3D box $(x,y,z,w,l,h,\theta)$ and a semantic category.
The BBox Autoencoder $\mathcal{E}_\textrm{box}$ processes slot-aligned 3D bounding box sequences by first converting each box into its eight 3D corners, applying a Fourier Positional Embedding (FPE), and concatenating a learnable class embedding. The resulting per-slot features are linearly projected into a hidden dimension and processed by a stack of spatiotemporal transformer blocks (ST-Transformer). Within these blocks, the model first performs frame-local spatial attention across object slots to capture cross-entity relationships, followed by temporal attention along the trajectory of each individual object slot to model temporal dynamics. Finally, a linear latent head predicts the posterior mean and log-variance for each slot, producing the bounding box latents $z_\textrm{box}$.

To accommodate scenes containing a variable number of objects, we allocate at most $N_{\max}$ slots and define a binary mask $M \in \{0,1\}^{N_{\max}}$ to pad invalid slots. The overall reconstruction objective is formulated as:
\begin{equation}
\begin{aligned}
\mathcal{L}_{\mathrm{BBox\text{-}AE}}
={}&\mathcal{L}_{\mathrm{corner}}
+\lambda_{\mathrm{yaw}} \mathcal{L}_{\mathrm{yaw}}
+\lambda_{\mathrm{cls}} \mathcal{L}_{\mathrm{cls}}\\
&+\lambda_{\mathrm{exist}} \mathcal{L}_{\mathrm{exist}}
+\lambda_{\mathrm{KL}} \mathcal{L}_{\mathrm{KL}}.
\end{aligned}
\end{equation}
The geometric and semantic losses are strictly evaluated over valid slots ($M_i=1$), which include an $\ell_1$ penalty on the 3D bounding box corners ($\mathcal{L}_{\mathrm{corner}} = \| P_i - \hat{P}_i \|_1$, where $P_i$ is the corner point), a continuous Euclidean loss in the $\textrm{SO}(2)$ embedding space for sine-cosine yaw regression ($\mathcal{L}_{\mathrm{yaw}} = \| [\sin\theta_i, \cos\theta_i]^\top - [\sin\hat{\theta}_i, \cos\hat{\theta}_i]^\top \|_2^2$, where $\theta$ is the yaw angle) to circumvent periodicity artifacts, and a cross-entropy loss for semantic categories ($\mathcal{L}_{\mathrm{cls}}$). Across all slots, $\mathcal{L}_{\mathrm{exist}}$ computes a binary cross-entropy loss for slot existence probability, and $\mathcal{L}_{\mathrm{KL}}$ provides the standard Kullback-Leibler divergence to regularize the latent distribution.

\section{Training and Implementation Details}
\label{sec:supp_multi_cam}

This section details the configuration and optimization strategies for our full $6$-camera surround-view training pipeline on the nuScenes dataset~\cite{caesar2020nuscenes}. We emphasize the spatial-temporal formulation, cross-view attention mechanism, and complex training curriculum.

\subsection{Data Preprocessing and Geometry Encoding}
Our model processes a temporal clip of $T=8$ frames across $N_C=6$ surrounding cameras per sample. The input images are resized to a resolution of $112 \times 168$. 
Given an input shape of $[B \times T, N_C, 3, H, W]$, the encoder downsamples the spatial resolution by a factor of $14$, yielding geometric latent tokens $z_{\mathrm{geo}} \in \mathbb{R}^{B \times N_C \times C \times T \times h \times w}$, where $C=1536$, $h=8$, and $w=12$.

\subsection{Condition Encoding with Wan-VAE}
The $6 \times 8$ input frames are not concatenated into a single $48$-frame video. Each camera stream is encoded independently as an $8$-frame video: the input $[B, T, N_C, 3, H, W]$ is reshaped into $B \times N_C$ videos of shape $[3, T, H, W]$, which the frozen Wan-VAE maps to condition latents of shape $[B, N_C, 16, 2, H/8, W/8]$.

\subsection{Camera Pose Encoding and Unified Spatial Grounding}
To spatially ground the geometric features, we inject explicit camera pose tokens into the latent space. For the $i$-th camera at the $t$-th frame, the continuous pose token $\mathcal{P}_{t}^{i}$ is constructed from both intrinsics (normalized by image width/height) and extrinsics (transformed relative to the ego coordinate system at $t=0$). The translation components are further normalized by a predefined scene radius factor $\rho=80$. 

To facilitate high-frequency spatial sensitivity, we apply a Fourier positional embedding $\gamma(\cdot)$ with $4$ frequency bands to the $6$-DoF pose vector:
\begin{equation}
    f_{\mathrm{pose}} = \mathrm{MLP} \bigl( \gamma(\mathcal{P}_{t}^{i}) \bigr).
\end{equation}
This camera pose embedding $f_{\mathrm{pose}}$ is subsequently added to the corresponding geometric tokens $z_{\mathrm{geo}}$ and concatenated as part of the conditional context for the diffusion backbone.

\subsection{Ring-Structured Cross-View Attention}
Instead of relying on computationally prohibitive full-attention across all cameras, we implement a memory-efficient ring-structured cross-view attention mechanism~\cite{gao2025magicdrive}. The $6$ cameras are modeled as a closed topological ring $(C_0 \rightarrow C_1 \rightarrow C_2 \rightarrow C_3 \rightarrow C_4 \rightarrow C_5 \rightarrow C_0)$. During the forward pass within each MMDiT block, the geometric tokens from camera $C_i$ only attend to their immediate spatial neighbors $C_{(i-1)\%6}$ and $C_{(i+1)\%6}$. It is important to note that this cross-view interaction is strictly applied to the dense geometry tokens $z_{\mathrm{geo}}$; the discrete bounding box tokens bypass the cross-view attention, are replicated across camera branches during inference, and are eventually averaged to produce the final layout predictions.




\subsection{Training Loss}

The total training loss combines the geometry diffusion objective, the layout latent flow-matching objective, and auxiliary layout supervision:
\begin{equation}
    \mathcal{L}_{\mathrm{total}}
    =
    \mathcal{L}_{\mathrm{geo}}
    +
    \lambda_\textrm{box}\mathcal{L}_{\mathrm{box}}
    +
    \mathcal{L}_{\mathrm{aux}},
\end{equation}
where $\mathcal{L}_{\mathrm{geo}}$ is the diffusion objective for the geometry branch, and $\mathcal{L}_{\mathrm{box}}$ is the velocity-matching loss for the BBox Autoencoder latent $z_{\mathrm{box}}$, evaluated only on valid object slots. We set $\lambda_\textrm{box}=5.0$ by default.

In addition, we supervise four auxiliary layout variables: object confidence $c$, anchor-relative center offset $\delta$, ground-plane velocity $v$, and object attribute $a$. For each auxiliary target $y_j$, where $j\in\{c,\delta,v,a\}$, we apply the same rectified-flow objective. Given Gaussian noise $\epsilon_j$ and noise level $\sigma$, we construct
\begin{equation}
    y_{j,\sigma} = (1-\sigma)y_j + \sigma \epsilon_j,
    \qquad
    \tilde{v}_j = \epsilon_j - y_j ,
\end{equation}
and train the corresponding prediction head $v_j$ with an MSE loss.

The confidence target is defined over all anchor slots, with $c=1$ for occupied slots and $c=-1$ for empty slots. For occupied slots, the center offset target is computed relative to the fixed anchor center:
\begin{equation}
\delta^{t,i}
=
\mathrm{clip}
\left(
\mathbf{x}_{\mathrm{box}}^{t,i}
-
\mathbf{x}_{\mathrm{anchor}}^{i}
,
-1,1
\right),
\end{equation}
The velocity target is the normalized ground-plane velocity, clipped to a bounded range:
\begin{equation}
v^{t,i}
=
\mathrm{clip}
\left(
\mathbf{u}^{t,i},
-3,3
\right).
\end{equation}
Finally, the attribute target $a^{t,i}$ is represented as a one-hot vector, and is supervised only when a valid attribute annotation is available\jr{.}

The auxiliary loss is therefore:
\begin{equation}
\mathcal{L}_{\mathrm{aux}}
=
\lambda_c\mathcal{L}_{c}
+
\lambda_\delta\mathcal{L}_{\delta}
+
\lambda_v\mathcal{L}_{v}
+
\lambda_a\mathcal{L}_{a}.
\end{equation}
In practice, we set $\lambda_c=\lambda_v=\lambda_a=1.0$ and $\lambda_\delta=2.0$.

\subsection{Training Curriculum and Flow Matching Hyperparameters}
Our unified MMDiT is initialized with the pretrained Wan2.1-1.3B backbone weights~\cite{wan2025wan}, leaving the DA3 encoder~\cite{lin2025da3} and the VAE components strictly frozen. We train the entire MMDiT trunk from the first epoch using AdamW optimizer. The flow-matching objective is simulated with $1000$ integration timesteps and a flow shift factor of $5.0$. 

To stabilize the optimization of high-frequency geometric details, we design a progressive three-stage noise curriculum:
\begin{itemize}
    \item \textbf{Phase I ($0\%$--$10\%$ progress):} High-noise sampling is disabled ($\sigma \leq 0.8$).
    \item \textbf{Phase II ($10\%$--$20\%$ progress):} Samples are drawn from a high-noise range of $[0.65, 0.9]$ with a $50\%$ probability.
    \item \textbf{Phase III ($20\%$--$100\%$ progress):} The high-noise sampling range is expanded to $[0.7, 1.0]$ with a $50\%$ probability.
\end{itemize}
Furthermore, to enhance robust unconditional generation capabilities, we apply a classifier-free guidance dropout of $p_{\mathrm{cfg}} = 0.1$ to the visual conditions.



\subsection{Model Configuration and Optimization Hyperparameters}

For the geometry autoencoder, the projected features are flattened into geometry tokens with embedding dimension $D_\textrm{g}=1536$. The geometry autoencoder is fine-tuned for $150$k iterations before joint training. For the BBox Autoencoder, object tracks are padded or truncated to at most $N_{\max}^{\mathrm{AE}}=100$ tracks per frame, and each track is mapped into a latent token of dimension $D_\textrm{b}=64$. The BBox Autoencoder is trained for $16$k iterations. We set the loss hyperparameters $\lambda_{\mathrm{exist}}=1.0$, $\lambda_{\mathrm{cls}}=0.5$, $\lambda_{\mathrm{yaw}}=1.0$, and $\lambda_{\mathrm{KL}}=10^{-4}$.

For joint diffusion, we instantiate $N_{\mathrm{anchor}}=1200$ fixed layout queries on a $40\times30$ BEV grid. The grid spans $x\in[-45,90]$\,m and $y\in[-60,60]$\,m, corresponding to a cell resolution of $3.375\times4.0$\,m. Each query is anchored at the center of its corresponding BEV cell with $z=0$, and the anchors are shared across all frames. The joint latent sequence is processed by a Multi-Modal Diffusion Transformer (MMDiT)~\cite{wan2025wan}, where the geometry and layout tokens are embedded by the proposed UPE module into a unified latent space for diffusion.

The unified MMDiT consists of 30 layers with hidden dimension $D=1536$ and 12 attention heads. For joint diffusion training, the model is trained on 8 NVIDIA H800 GPUs for $1500$k iterations using the AdamW optimizer with an initial learning rate of $1\times10^{-4}$, followed by cosine decay to $1\times10^{-5}$. We use a total batch size of 8, with one video clip per GPU and no gradient accumulation. Gradient clipping with a maximum norm of 1.0 is applied for stability.

\subsection{Decoding and Post-Processing at Inference}
We use $50$ rectified-flow denoising steps at inference. Since the learned frame embeddings and box tensors are fixed to $T{=}8$, sequences longer than $8$ frames are processed with overlapping $8$-frame sliding windows.
At inference time, the generated geometry latent is denormalized and propagated through the DPT decoder to obtain 3D Gaussian representations, while layout tokens are denormalized and decoded by the frozen BBox-AE into 3D bounding boxes, class logits, and existence logits. Box centers are recovered from the predicted anchor-relative offsets $\delta$, and final detections are selected by confidence thresholding and per-frame 3D non-maximum suppression.

\section{Efficiency}
On a single H800 GPU, USR-Drive processes one $6$-camera $\times$ $8$-frame clip in $45.2$\,s with $58.5$\,GB peak memory, dominated by the $50$-step iterative denoising, while the underlying DA3 encoder alone takes $1.3$\,s and $46.6$\,GB. Unlike per-scene optimization methods that require hours of test-time optimization per scene, our model is fully feed-forward apart from the denoising loop, and the training cost is amortized across all scenes. USR-Drive targets offline unified scene-state reconstruction and perception rather than real-time onboard deployment; distilling the multi-step denoiser into a few-step causal model is a promising direction for future work.

\section{More Ablation Studies}

\subsection{Ablation on Auxiliary Layout Variables}
We ablate how to incorporate auxiliary layout variables, including object confidence $c$, anchor-relative center offset $\delta$, velocity $v$, and dynamic/static attribute $a$. One alternative is to encode these variables directly into the BBox Autoencoder and reconstruct them together with the box representation. However, this couples the autoencoder to the fixed-anchor denoising design: confidence is defined over both occupied and empty anchor slots, and $\delta$ depends on the anchor centers used by the diffusion model. Any change to these targets or the anchor design would then require retraining the autoencoder. In addition, velocity and attribute labels can be sparse or unavailable for some boxes, which is more naturally handled with masked auxiliary diffusion heads. Therefore, we keep the BBox Autoencoder focused on the core box latent and add $c$, $\delta$, $v$, and $a$ as separate denoising branches in the layout diffusion model. This preserves the pretrained autoencoder, keeps the auxiliary variables explicit and controllable, and enables clean ablations. Results are reported in Tab.~\ref{tab:supp_auxiliary}.

\begin{table}[t]
\centering
\small
\setlength{\tabcolsep}{3pt}

\begin{tabular}{lccc}
\toprule
Auxiliary variable placement & mAP$\uparrow$ & mATE$\downarrow$ & PSNR$\uparrow$ \\
\midrule
No auxiliary & 0.223 & 0.801 & 27.49 \\
Inside BBox Autoencoder & 0.445 & 0.679 & 27.31 \\
Separate Learnable Tokens \textbf{(Ours)} & \textbf{0.552} & \textbf{0.525} & \textbf{27.55} \\
\bottomrule
\end{tabular}
\caption{Ablation on auxiliary layout tokens.}
\label{tab:supp_auxiliary}
\end{table}

\subsection{Ablation on Training Parameters}
We further ablate two key training hyperparameters: the bounding-box loss weight $\lambda_{\textrm{box}}$ and the high-noise curriculum, with all other settings kept unchanged.

As shown in Tab.~\ref{tab:ablation_training}, $\lambda_{\textrm{box}}$ balances dense geometry reconstruction and sparse layout prediction. A small weight improves reconstruction but hurts detection, while an overly large weight over-emphasizes boxes and degrades PSNR/SSIM. We use $\lambda_{\textrm{box}}=5.0$ by default, which gives the best overall trade-off.

For the high-noise curriculum, our default setting disables high-noise sampling in the first $10\%$ of training, enables it with probability $0.5$ from noise range $[0.65,0.9]$ during $10\%$--$20\%$, and expands the range to $[0.7,1.0]$ afterwards. Removing this curriculum destabilizes early training and degrades both reconstruction and detection, indicating that gradually increasing denoising difficulty helps the model learn local geometry before global layout.

\begin{table}[t]
    \centering
    \small
  
    \begin{tabular}{l|cc|cc}
            \toprule
            \textbf{Configuration} & \textbf{PSNR$\uparrow$} & \textbf{SSIM$\uparrow$} & \textbf{mAP$\uparrow$} & \textbf{mATE$\downarrow$} \\
            \midrule
            $\lambda_{\textrm{box}} = 1.0$ & 27.42 & 0.849 & 0.449 & 0.624 \\
            $\lambda_{\textrm{box}} = 3.0$ & 27.50 & 0.851 & 0.491 & 0.575 \\
            $\lambda_{\textrm{box}} = 7.0$ & 27.20 & 0.841 & 0.521 & 0.540 \\
            $\lambda_{\textrm{box}} = 10.0$ & 26.38 & 0.812 & 0.516 & 0.548 \\ \midrule
            $\lambda_{\textrm{box}} = 5.0$, w/o Curr. & 26.95 & 0.831 & 0.473 & 0.585 \\
            \midrule
            \textbf{$\lambda_{\textrm{box}}=5.0$, w/ Curr.} & \textbf{27.55} & \textbf{0.853} & \textbf{0.552} & \textbf{0.525} \\
            \bottomrule
    \end{tabular}
    \caption{Ablation on training hyperparameters. We evaluate the bounding-box loss weight $\lambda_{\textrm{box}}$ and the high-noise curriculum. The default setting achieves the best balance between geometric fidelity and layout accuracy.}
    \label{tab:ablation_training}
\end{table}

\subsection{Ablation on DiT Hidden Dimension}
We ablate the MMDiT hidden dimension $D$ while keeping all other settings unchanged. As shown in Tab.~\ref{tab:ablation_dimension}, smaller $D$ reduces cost but creates a bottleneck for 1536-channel geometry features and breaks compatibility with pretrained Wan2.1-1.3B weights. Larger $D$ (e.g., 2048) greatly increases parameters and training cost with only marginal gains, while also losing pretrained initialization. Thus, we use $D=1536$ as the best trade-off between capacity, efficiency, and architectural compatibility.

\begin{table}[t]
    \centering
    \small
    \setlength{\tabcolsep}{2pt}
    
    \begin{tabular}{l|c|ccc}
        \toprule
        \textbf{Feature Dim.} & \textbf{Params (B)} & \textbf{PSNR$\uparrow$} & \textbf{SSIM$\uparrow$} & \textbf{LPIPS$\downarrow$} \\
        \midrule
        768  & 0.356 & 25.84 & 0.792 & 0.138 \\
        1024 & 0.631 & 26.68 & 0.825 & 0.104 \\
        1536 (Default) & 1.418 & \textbf{27.55} & \textbf{0.853} & \textbf{0.076} \\
        2048 & 2.517 & 27.48 & 0.849 & 0.080 \\
        \bottomrule
    \end{tabular}
    \caption{Ablation on MMDiT hidden dimension. The default 1536-dimensional setting achieves the best trade-off between reconstruction quality and model size.}
    \label{tab:ablation_dimension}
\end{table}

\section{Qualitative Results}

\subsection{Reconstruction Results on nuScenes}
\begin{figure*}[t]
    \centering
    \includegraphics[width=0.94\linewidth]{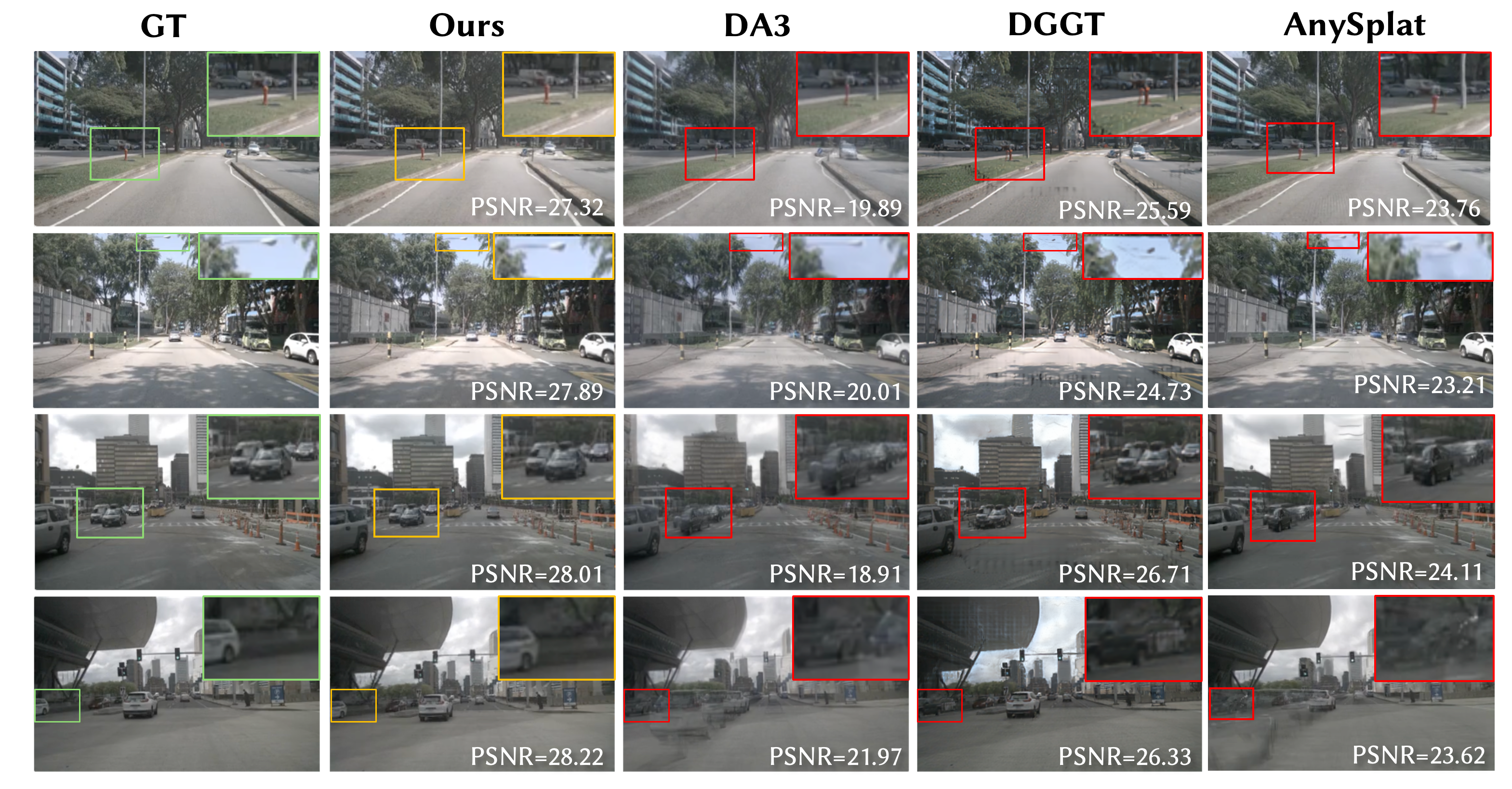}
    \caption{\textbf{Visualization of reconstruction results.} Compared to baseline methods, our approach reveals better appearance quality on both static scenes (top 2 rows) and dynamic objects (bottom 2 rows).}
    \label{fig:exp1}
\end{figure*}

Fig.~\ref{fig:exp1} presents qualitative reconstruction results on the nuScenes dataset. Compared to baseline methods, our approach reveals better appearance quality on both static scenes (top 2 rows) and dynamic objects (bottom 2 rows). In particular, our method recovers sharper scene details, cleaner object boundaries, and more coherent dynamic content under challenging driving scenarios, benefiting from the unified latent modeling of dense geometry and sparse object layout.

\subsection{3D Detection Results on nuScenes}
\begin{figure*}[t]
    \centering
    \includegraphics[width=0.92\linewidth]{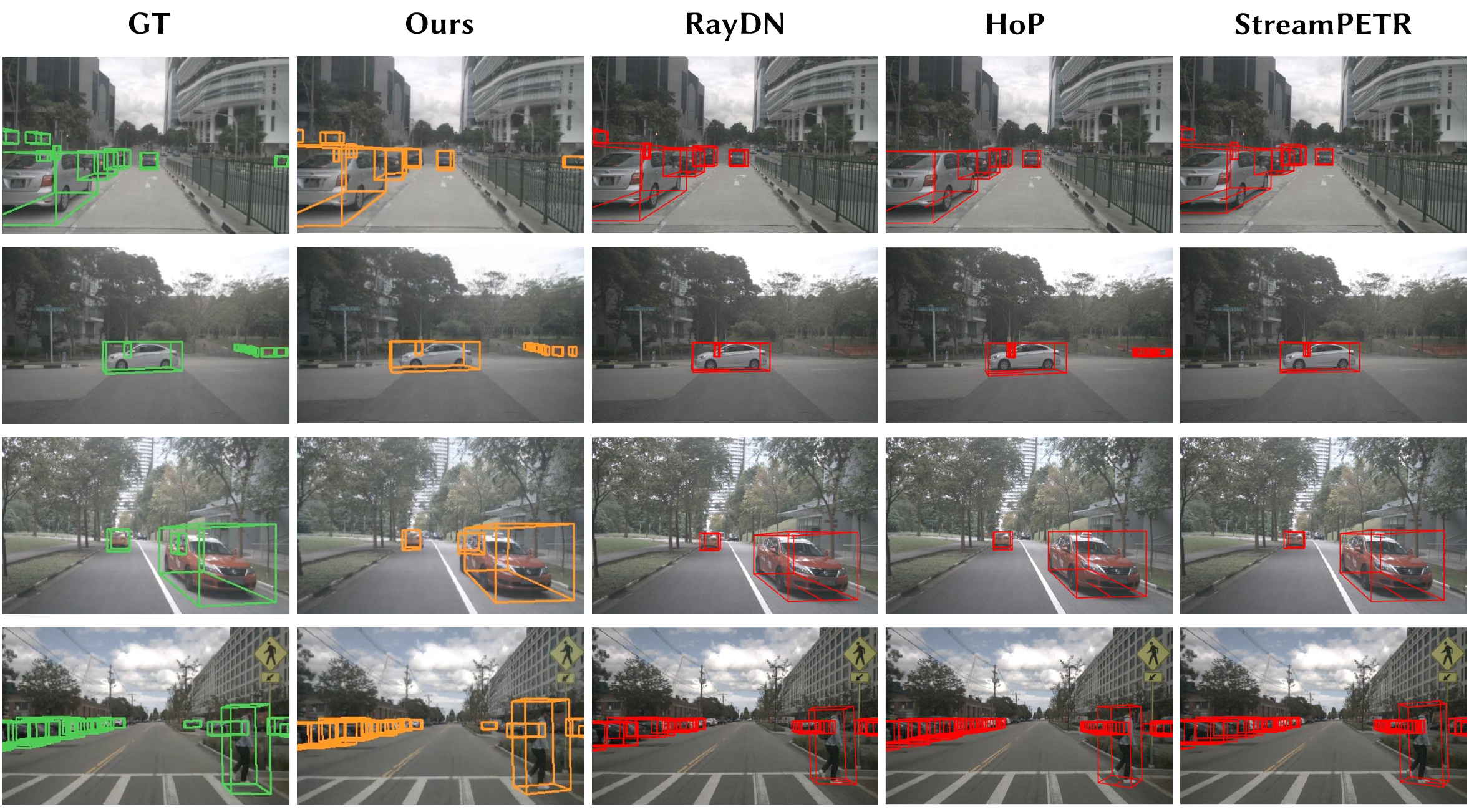}
    \caption{\textbf{Visualization of 3D detection results.} While previous methods suffer from missed or inaccurate detection, our method achieves better results considering completeness and accuracy.}
    \label{fig:exp2}
\end{figure*}

Fig.~\ref{fig:exp2} visualizes 3D detection results on the nuScenes dataset. While previous methods suffer from missed or inaccurate detection, our method achieves better results in terms of both completeness and accuracy. This indicates that the joint generation paradigm benefits from the information of the geometric branch, which provides structural context and spatial grounding for object-level detection.

\subsection{Visualization of Ablation Studies}
\begin{figure*}[t]
    \centering
    \includegraphics[width=0.92\linewidth]{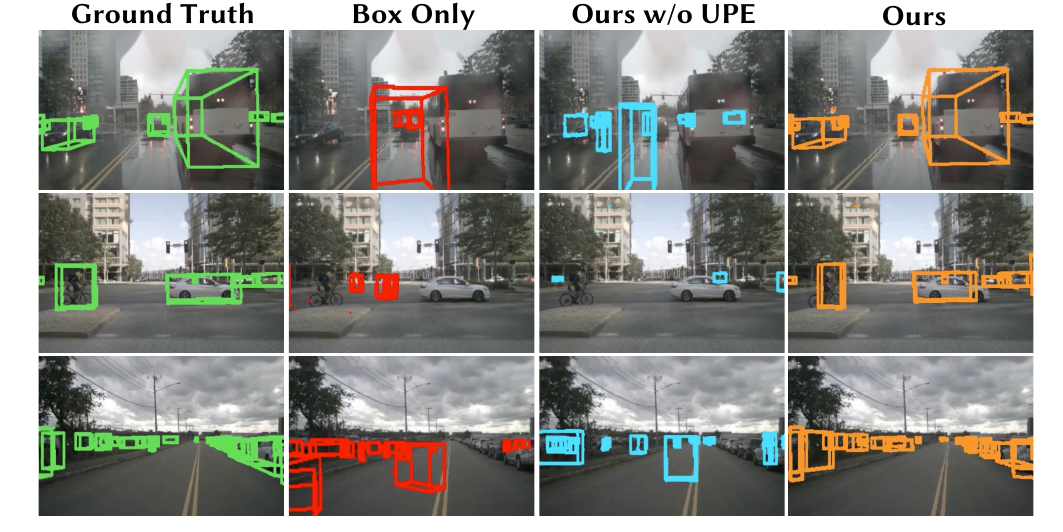}
    \caption{\textbf{Visualization of Ablation Studies.} The complete model provides the best reconstruction effect and detection quality.}
    \label{fig:ablation1}
\end{figure*}

Fig.~\ref{fig:ablation1} visualizes the ablation variants reported in the main paper. The geometry-only variant produces reasonable appearances but lacks object-level structure, while removing UPE leads to visible misaligned objects and implausible box placements. The complete model provides the best reconstruction quality and detection accuracy, confirming the necessity of both the co-denoising design and the unified positional encoding.

\subsection{Zero-Shot Results on VKitti}
\begin{figure*}[t]
    \centering
    \includegraphics[width=0.9\linewidth]{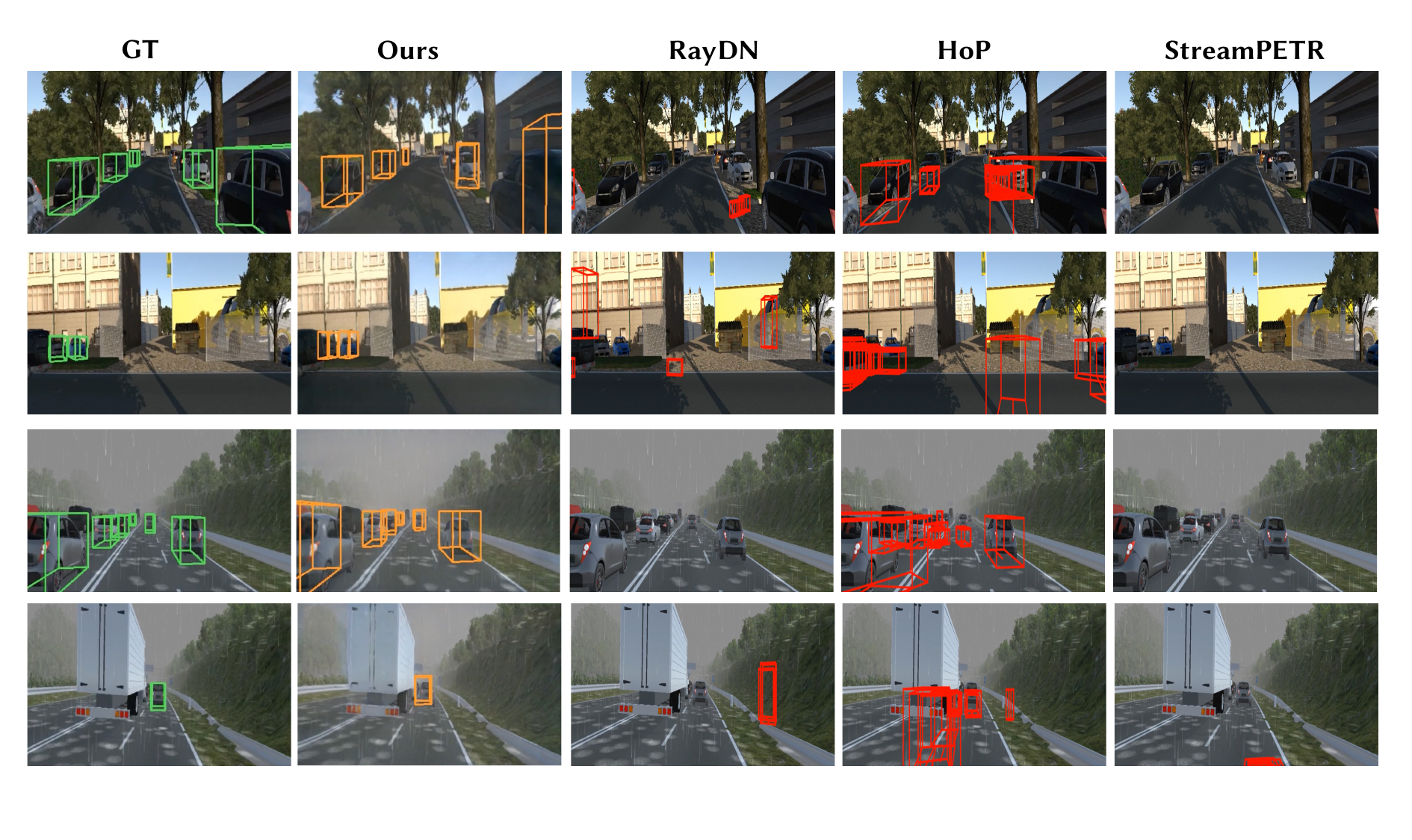}
    \caption{\textbf{Visualization of 3D detection results on VKitti dataset.} While previous methods suffer from missed or inaccurate detection, our method achieves better results considering completeness and accuracy.
}
    \label{fig:vkitti_detection}
\end{figure*}

\begin{figure*}[t]
    \centering
    \includegraphics[width=0.9\linewidth]{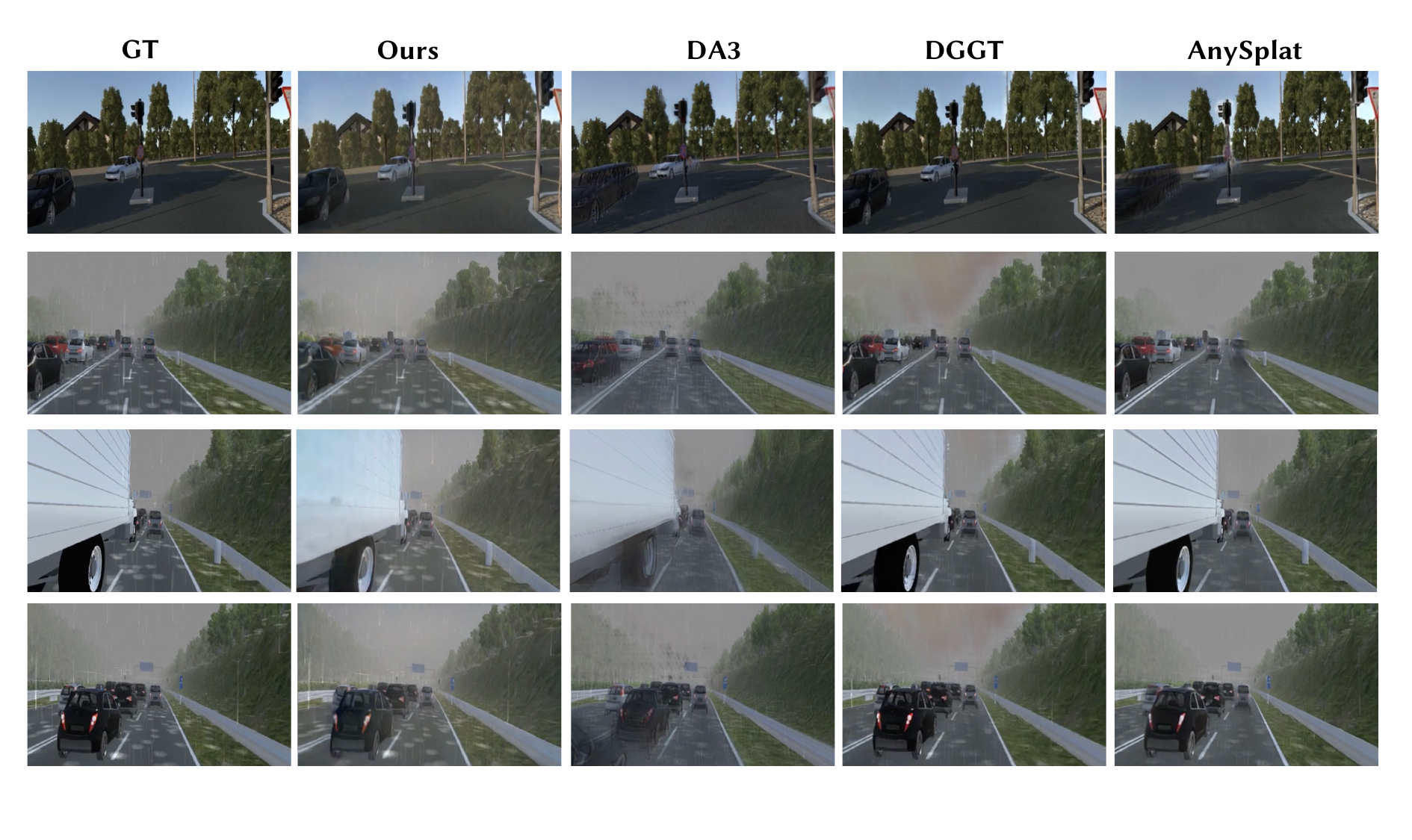}
    \caption{\textbf{Visualization of reconstruction results on VKitti dataset.} Compared to baseline methods, our approach reveals better appearance quality.}
    \label{fig:vkitti_recon}
\end{figure*}

Fig.~\ref{fig:vkitti_detection} and Fig.~\ref{fig:vkitti_recon} show zero-shot 3D detection and reconstruction results on the VKitti dataset, respectively. Without any fine-tuning, our method transfers well to the unseen synthetic domain, producing tightly grounded bounding boxes and high-fidelity reconstructions, whereas baseline methods suffer from missed detections or degraded appearance quality.

\subsection{Depth Estimation}
\begin{figure*}[t]
    \centering
    \includegraphics[width=\linewidth]{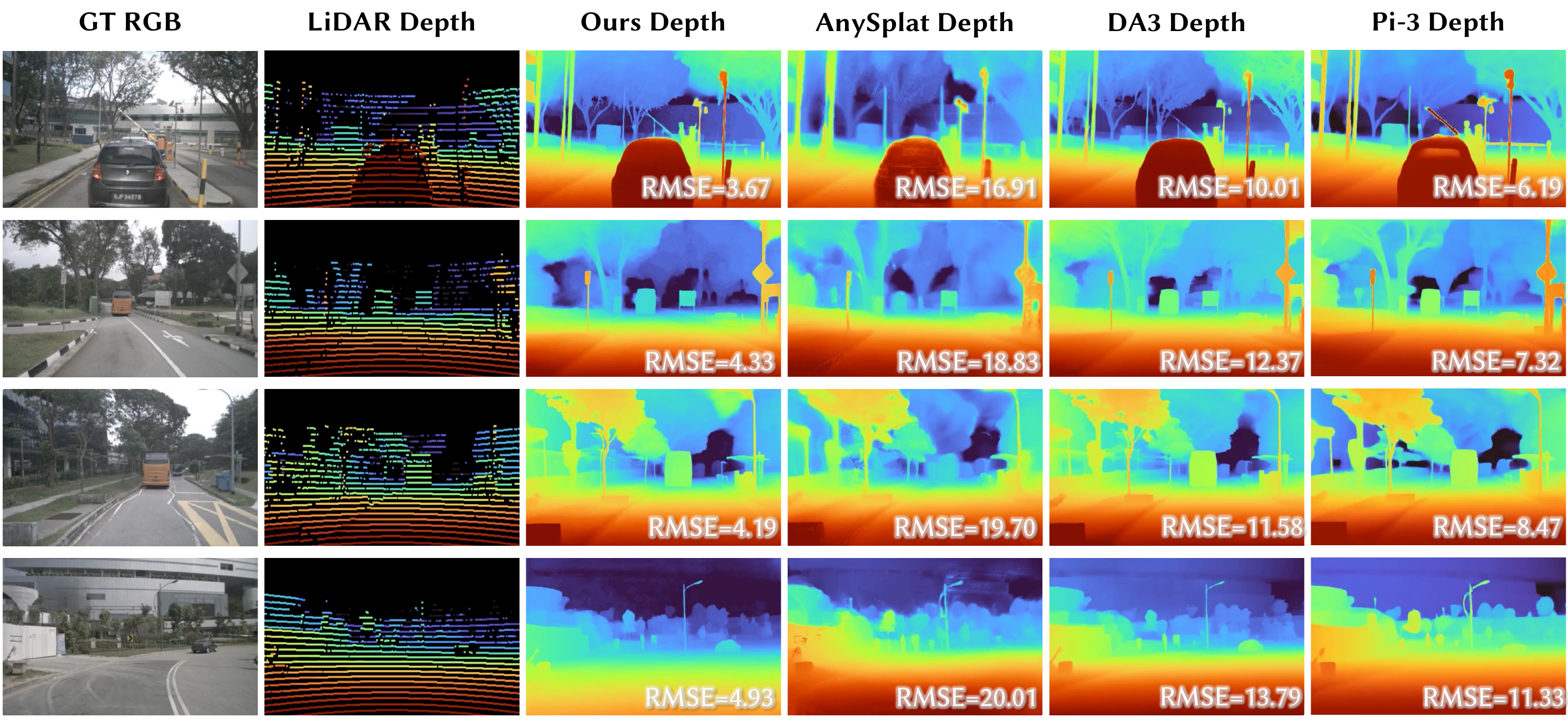}
    \caption{\textbf{Visualization of Depth results.} Our method achieves best depth estimation on both qualitative and quantitative results.
}
    \label{fig:exp3}
\end{figure*}

Fig.~\ref{fig:exp3} compares the depth maps derived from the reconstructed 3D Gaussians. Our method achieves the best depth estimation in both qualitative and quantitative evaluations, further verifying that the unified latent space preserves accurate metric scene structure.

\end{document}